\documentclass[runningheads]{llncs}
\usepackage[hidelinks]{hyperref}
\usepackage{orcidlink}
\usepackage[T1]{fontenc}
\usepackage{graphicx}

\usepackage{academicons}
\usepackage{orcidlink}
\usepackage{tabularx}
\usepackage{array}
\usepackage{tikz}
\usepackage{multirow}
\usetikzlibrary{trees}
\usepackage{amsmath}

\usepackage[T5]{fontenc}
\usepackage{balance}
\usepackage[utf8]{inputenc}
\DeclareTextSymbolDefault{\DH}{T1}
\usepackage{lmodern} 
\usepackage{amssymb}
\usepackage{footmisc}

\def\BibTeX{{\rm B\ker†n-.05em{\sc i\kern-.025em b}\kern-.08em
    T\kern-.1667em\lower.7ex\hbox{E}\kern-.125emX}}
\begin{document}
\title{MasHeNe: A Benchmark for Head and Neck CT Mass Segmentation using Window-Enhanced Mamba with Frequency-Domain Integration}
\titlerunning{MasHeNe: A Benchmark for Head and Neck CT Mass Segmentation}
%
\author{Thao Thi Phuong Dao\inst{1,2,3}\orcidlink{0000-0002-0109-1114} \and
Tan-Cong Nguyen\inst{1,2,4}\orcidlink{0000-0001-8834-8092} \and
Nguyen Chi Thanh\inst{3}\orcidlink{0009-0002-8108-1502} \and
Truong Hoang Viet\inst{3}\orcidlink{0009-0007-9433-0017} \and
Trong-Le Do\inst{1,2}\orcidlink{0000-0002-2906-0360} \and
Mai-Khiem Tran\inst{1,2}\orcidlink{0000-0001-5460-0229} \and
Minh-Khoi Pham\inst{5}\orcidlink{0000-0003-3211-9076} \and 
Trung-Nghia Le\inst{1,2}\orcidlink{0000-0002-7363-2610} \and
Minh-Triet Tran\inst{1,2,7}\orcidlink{0000-0003-3046-3041}\thanks{Corresponding authors: Minh-Triet Tran (tmtriet@fit.hcmus.edu.vn) and Thanh Dinh Le (ledinhthanhvmc@yahoo.com.vn)}\and
Thanh Dinh Le\inst{3,6,2}\orcidlink{0009-0009-3153-085X}$^{\ast}$}

\authorrunning{Thao TP Dao, Tan-Cong Nguyen et al.}
%
\institute{University of Science, VNU-HCM, Ho Chi Minh City, Vietnam \and
Vietnam National University,  Ho Chi Minh City, Vietnam \and
Thong Nhat Hospital, Ho Chi Minh City, Vietnam  \and University of Social Sciences and Humanities, VNU-HCM, Ho Chi Minh City, Vietnam \and Dublin City University, Dublin, Ireland \and University of Health Sciences, VNU-HCM, Ho Chi Minh City, Vietnam  \and John von Neumann Institute, VNU-HCM, Ho Chi Minh City, Vietnam}
\pagestyle{empty}
\maketitle              
\begin{abstract}
Head and neck masses are space-occupying lesions that can compress the airway and esophagus and may affect nerves and blood vessels. Available public datasets primarily focus on malignant lesions and often overlook other space-occupying conditions in this region. To address this gap, we introduce MasHeNe, an initial dataset of 3,779 contrast-enhanced CT slices that includes both tumors and cysts with pixel-level annotations. We also establish a benchmark using standard segmentation baselines and report common metrics to enable fair comparison.
In addition, we propose the Windowing-Enhanced Mamba with Frequency integration (WEMF) model. WEMF applies tri-window enhancement to enrich the input appearance before feature extraction. It further uses multi-frequency attention to fuse information across skip connections within a U-shaped Mamba backbone. On MasHeNe, WEMF attains the best performance among evaluated methods, with a Dice of 70.45 \%, IoU of 66.89 \%, NSD of 72.33 \%, and HD95 of 5.12 mm. This model indicates stable and strong results on this challenging task.
MasHeNe provides a benchmark for head-and-neck mass segmentation beyond malignancy-only datasets. The observed error patterns also suggest that this task remains challenging and requires further research. Our dataset and code will be made publicly available at \url{https://github.com/drthaodao3101/MasHeNe.git}.
\end{abstract}
\section{Introduction}
\label{sec:int}

Head and neck masses are common pathological entities that present as space-occupying lesions displacing or compressing adjacent structures \cite{pynnonen2017clinical}. Consequences include airway narrowing, esophageal obstruction, involving nerves and major vessels. These effects threaten breathing, swallowing, voice, or cerebral perfusion. The causes of head and neck masses are diverse. It includes congenital anomalies, inflammatory or infectious conditions, benign or malignant neoplasms, and metabolic or autoimmune disorders \cite{schwetschenau2002adult}.
Among imaging modalities, contrast-enhanced computed tomography (CECT) is the preferred modality for evaluating neck masses. It is fast, widely available, and offers high spatial resolution \cite{schwetschenau2002adult}. It effectively aids in distinguishing cystic from solid lesions, determining disease extent, and guiding biopsy. Therefore, precise mass segmentation on CECT supports volumetry, surgical planning, and treatment monitoring. 

However, building robust segmentation models for the head and neck remains challenging. The regional anatomy is dense and complex, with many small structures packed into a narrow field. Lesions vary in size, shape, and patterns. Besides, low contrast around lesion borders further obscures boundaries. Moreover, data limitations compound these challenges. Many public datasets, such as StructSeg \cite{li2020automatic}, RADCURE \cite{welch2024radcure}, and SegRap2023 \cite{luo2025segrap2023}, focus on malignant disease and lack coverage of other space-occupying conditions, such as cysts and benign tumors. This malignancy-focused bias restricts the range of appearances and pathologies seen during training. In real clinical practice, the case mix includes both malignant and non-malignant masses. A model trained only on malignant lesions may overfit to enhancement cues and fail on non-enhancing.

To address these gaps, we introduce MasHeNe, a new dataset for head and neck mass segmentation in contrast-enhanced CT. MasHeNe with pixel-level annotations comprises 65 cases with a total of 3,779 slices, covering tumors and cysts. Alongside the dataset, we establish a set of baseline models. This benchmark setting enables future studies to reproduce our results, develop algorithms, and measure progress under consistent conditions.
We further propose a Windowing-Enhanced Mamba integrating the Frequency domain (WEMF) model to exploit complementary CT appearances. The method applies a tri-windowing supplement that enriches the input by composing three clinically meaningful windows. These windows expose different intensity ranges and edge profiles, which improve the visibility of both low-contrast boundaries and high-frequency details. Within a U-shaped Mamba backbone, we introduce multi-frequency attention fusing information across frequency representations along skip connections. This design enables the network to preserve information across frequency bands and enhance it for the decoder branch, yielding improved boundary detection and robustness to appearance changes. 
On MasHeNe, WEMF achieves best performance among methods, with Dice, IoU, NSD, and HD95 of 70.45 \%, 66.89 \%, 72.33 \%, and 5.12 mm, respectively. These values indicate stable results on a challenging task and show the benefit of combining multi-window inputs with frequency-aware fusion. Our main contributions are as follows:

\begin{itemize}
  \item \textbf{MasHeNe dataset.} We propose MasHeNe, a novel head and neck contrast-enhanced CT dataset with 65 cases.
  \item \textbf{Benchmark for MasHeNe dataset.} We establish a benchmark with baseline models across CNN, Transformer, and Mamba families, together with a unified evaluation protocol to enable fair comparison.
  \item \textbf{WEMF model.} We propose the Windowing-Enhanced Mamba integrating the Frequency domain (WEMF) model, which applies tri-windowing enhancement and uses multi-frequency attention across the skip connections.
\end{itemize}

\vspace{-5mm}
\section{Related works}
\label{sec:rel}

\subsection{Available CT Image Dataset for Head and Neck Segmentation}

Public datasets have enabled advanced progress in head and neck CT segmentation, particularly for oncology workflows that delineate tumors and organs–at–risk (OARs). The Cancer Imaging Archive (TCIA) \cite{clark2013cancer} is the primary source for many widely used resources. From TCIA, several benchmarks were curated, including Public Domain Database for Computational Anatomy (PDDCA), which provides expert OAR labels from the Head–Neck Cetuximab (HNC) cohort \cite{raudaschl2017evaluation}, the Test \& Validation Radiotherapy CT Planning set aggregates cases from HNC and TCGA–HNSC with multi–OAR annotations \cite{nikolov2021clinically,bejarano2019longitudinal}, and UaNet expanding the labeled cohort using HNC and Head-Neck-PET-CT (HNPETCT), increasing both case count and OAR coverage \cite{tang2019clinically,vallieres2017radiomics}. 

Beyond TCIA–derived resources, other datasets broaden the landscape. StructSeg releases CECTs with a standardized set of OAR labels \cite{li2020automatic}. Paired–modality collections such as RT–MRI and HaN–Seg offer CT and MRI pairs with comprehensive OAR annotations to support cross–modal learning and domain adaptation \cite{head2019prospective,podobnik2023han}. Large–scale cohorts like RADCURE include thousands of patients with tumors, nodal disease, and multiple OARs, enabling robust model training and validation \cite{welch2024radcure}. For disease–specific tasks, SegRap2023 provides paired contrast and non–contrast CTs with OARs and gross tumor volumes in nasopharyngeal carcinoma \cite{luo2025segrap2023}.


\subsection{Window setting in CT scans}

Window setting, including window width and level, is a basic step in CT image analysis \cite{hoang2010ct}. It maps the wide Hounsfield Unit (HU) range to a displayable gray scale so that the contrast of lesion and background becomes visible to the human eye. Window width (WW) controls how many HU values are included in the gray scale. Values that are below and above this range appear black and white. Window level (WL) selects the center HU of that range. 

 Recent studies have adapted CT windowing for model training. Huo et al. \cite{huo2019stochastic} propose stochastic tissue window normalization, which randomly perturbs the window level and width around a soft-tissue setting during training to normalize CT intensities and improve a 2D U-Net’s generalization. Apivanichkul et al. \cite{apivanichkul2021performance} apply custom window-leveling as preprocessing before training a U-Net for organ segmentation, yielding higher accuracy than standard windows. Thariat et al. \cite{thariat2025auto} modify nnU-Net with a triple-window input to avoid the compression caused by percentile-based normalization. The added windows enhance soft-tissue contrast, especially in the pharyngeal mucosa. Østmo et al. \cite{ostmo2023view} introduce window shifting, a CT-coherent augmentation that samples window levels around a base view. Lin et al. \cite{lin2025multi} learn from raw, lung-window, and mediastinal-window images for nodule segmentation, using a probabilistic Attention U-Net with an Uncertainty Indication Module and a Feature Filter to build a Multi-Window Mask. Overall, these works support windowing variants, adaptive window selection, and window-aware augmentations for CT segmentation of organs or lesions.
 
\vspace{-2mm}
\subsection{Baselines for Medical Image Segmentation}

Convolutional neural networks (CNNs) with demonstrated results show their effectiveness in medical image segmentation. U-Net \cite{ronneberger2015u} and its variants, such as U-Net++ \cite{zhou2018unet++}, Attention U-Net \cite{oktay2018attention}, Residual U-Net \cite{alom2018recurrent}, use an encoder–decoder with skip connections to combine high-level semantics and fine spatial details. These models are data-efficient, easy to train, and run fast on standard GPUs. Common improvements include multi-scale feature fusion, astrous convolutions, deep supervision, and lightweight decoders \cite{yuan2024medical,das2024acdssnet,lee2015deeply,feng2024mlu}.  CNN baselines are often strong for well-defined boundaries and limited domain shift, but they may struggle with long-range context and variable contrast. Transformers \cite{transformer} introduce self–attention to capture long-range dependencies and global context. In medical imaging, hybrid designs, such as TransUNet \cite{chen2021transunet}, combine CNN encoders with Transformer blocks, while fully Transformer architectures, such as UNETR \cite{hatamizadeh2022unetr}, Swin-UNETR \cite{hatamizadeh2021swin}, SegFormer \cite{xie2021segformer}, style decoders process tokens at multiple scales. Self–attention helps delineate ambiguous borders and model non-local relationships between tissues. However, pure Transformers can be memory-intensive and may require larger datasets or strong regularization. 

Until recent, State Space Models (SSMs) such as Mamba \cite{gu2023mamba} provide an alternative to heavy self–attention by using selective state transitions with linear-time complexity. Recent works apply Mamba blocks or Visual State Space (VSS) blocks to U-shaped segmentation backbones in the encoder and decoder such as VM-UNet \cite{ruan2024vm} and U-Mamba \cite{ma2024u}. Besides, Swin U-Mamba \cite{liu2024swin} employs VSS blocks in the encoder and inserts a residual convolutional block on each skip to refine features before fusion.
These models aim to capture long-range dependencies with lower memory cost and better throughput than Transformers, while preserving locality similar to CNNs. 

\vspace{-2mm}
\section{MasHeNe Dataset}
\label{sec:data}

\subsection{Data Collection}

In this study, we collected contrast-enhanced CT scans of head and neck masses from 65 cases with 3,779 slices. The cases were randomly divided into a training set with 50 cases (2,931 slices), a validation set with 5 cases (278 slices), and a test set with 10 cases (570 slices). This dataset consisted of 29 cystic and 36 tumor cases. We excluded a number of poor-quality scans from the study. These included images blurry from patient movement, contained metal-induced artifacts in dental works, or did not fully cover the required anatomical area.

All CT scans were acquired on a Philips Brilliance iCT SP 128 scanner (Philips Healthcare, Best, Netherlands). Standardized protocols were used with parameters, including tube voltage 120 kVp, automatic tube current modulation 200 to 300 mAs, slice thickness 1.5 to 3.0 mm, and in-plane resolution 0.5 × 0.5 mm with a 512 × 512 matrix. All imaging data were collected in Digital Imaging and Communications in Medicine (DICOM) format. Lesions on each CT series were manually annotated in 3D Slicer \cite{pieper20043d}to precisely define their boundaries. The resulting segmentation masks were then converted to the Nearly Raw Raster Data (NRRD) format for downstream processing and analysis.

To protect patient privacy, all data were fully anonymized before analysis in accordance with Thong Nhat Hospital ethical guidelines and the principles of the Declaration of Helsinki. The study protocol received formal approval from the Institutional Ethics Committee of Thong Nhat Hospital, Approval No. 70/2024/CN-BVTN-H\DJ{}\DJ{}\DJ{}.

\vspace{-2mm}
\subsection{Data Annotation}
\subsubsection{Mask Define.}

On CECT scans, the first group of findings is tumors. These appear as distinct masses. Their internal appearance depends on what they are made of, like solid tissue, fluid, or a mix. After contrast injection, the way they enhance helps suggest the tumor's tissue type, areas of dead tissue, or bleeding inside it. It also shows their relationship to blood vessels. Features that suggest a benign tumor include a smooth, well-defined border, often with a capsule. The inside of the mass is relatively uniform. It doesn't invade surrounding tissues but just pushes them aside. It enhances evenly with contrast, does not invade blood vessels, and has no abnormal lymph nodes in the neck. 
In contrast, features that suggest a malignant, called cancerous, tumor include an irregular border and invasion into nearby tissues. The inside of the mass is not uniform. It may invade blood vessels or nerves, and there may be abnormal neck lymph nodes. 

The second group in our dataset is cysts. A typical cyst has a smooth, thin, and well-defined wall. Its inside is uniform and has a density close to water, ranging from 0 to 20 HU. After contrast injection, the center of the mass does not enhance. Sometimes the thin wall may enhance slightly or not at all. A cystic lesion does not invade nearby tissues; instead, it simply displaces them.

\subsubsection{Annotation Progress.}

We used a three-stage workflow to produce accurate and consistent lesion masks. The annotation was carried out by clinicians with expertise in head-and-neck imaging, including otolaryngologists and radiologists. All work was conducted according to a written protocol to standardize procedures and minimize subjectivity.

\begin{itemize}
    \item \textbf{Stage 1: Independent annotation}. The dataset was assigned to two clinicians who worked independently. For each CT slice, they manually labelled the lesion at the pixel level using dedicated medical-imaging software. Each annotator segmented the full extent of the lesions for every case and saved the masks for later comparison.

    \item \textbf{Stage 2: Consensus review}. Cases showing any disagreement were jointly re-examined by the two annotators. They discussed ambiguous boundaries, negotiated differences, and produced a single consensus mask per case. This step was designed to reduce inter-annotator variability and improve label reliability. Any case that could not be resolved by consensus was escalated.

    \item \textbf{Stage 3: Expert adjudication and quality control}. A senior head-and-neck expert independently reviewed all consensus masks and the escalated disputes. The expert verified anatomical correctness, checked cross-slice continuity, and made final edits when needed. The adjudicated masks constitute the reference ground truth used in our study.
\end{itemize}

\vspace{-6mm}
\section{Method}
\label{sec:methods}

\subsection{Overview of the WEMF Architecture}
\label{subsec:overview}

\begin{figure*}[t]
  \centering
    \includegraphics[width=0.9\textwidth,trim={3cm 0cm 0cm 0cm}, clip]{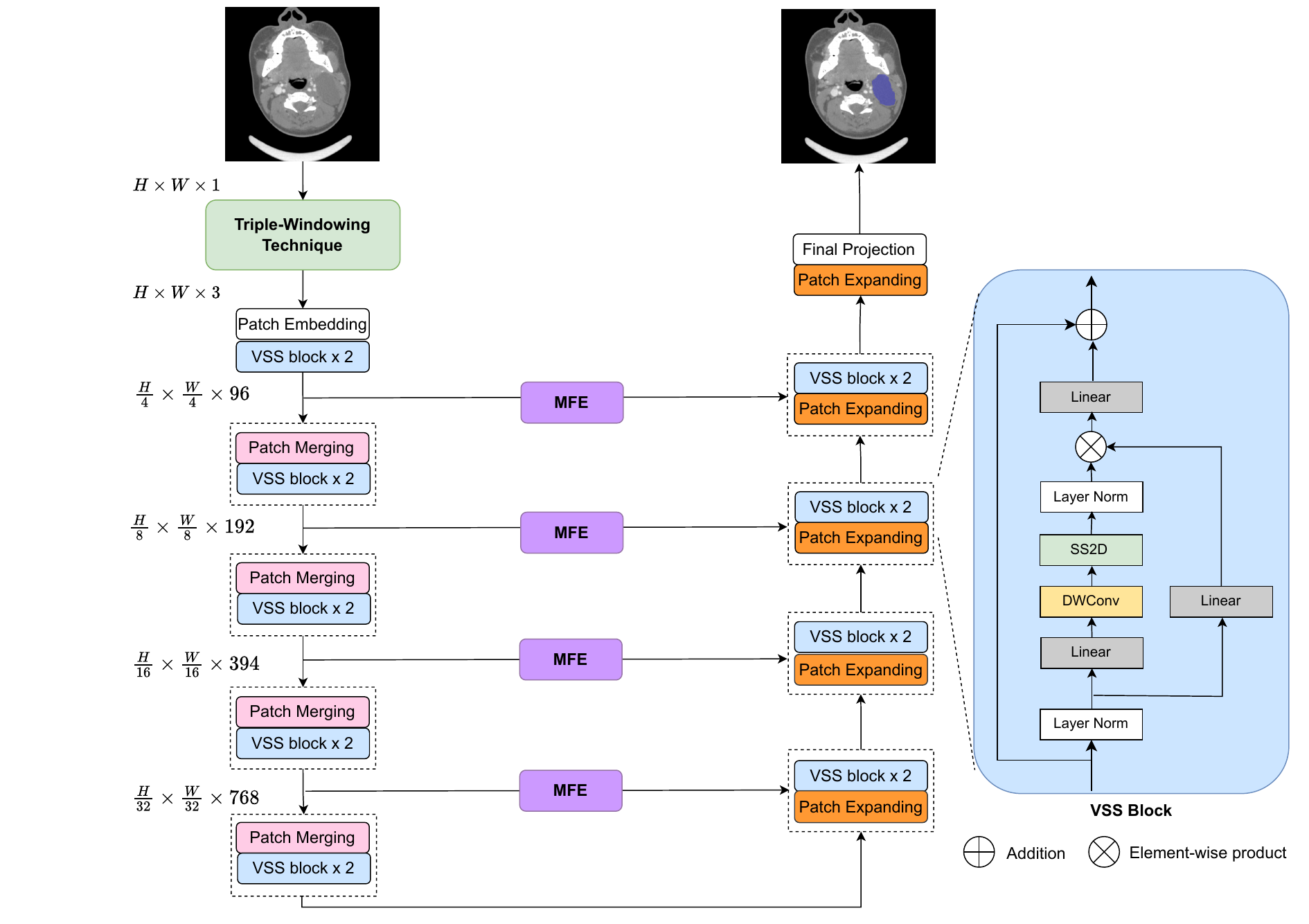}
    \vspace{-5 mm}
    \caption{Overview architecture of our WEMF model}
    \label{fig:model}
    \vspace{-6 mm}
\end{figure*}

Our network follows a U\mbox{-}shaped encoder–decoder built on VSS blocks as Fig.~\ref{fig:model}. At the input, a Triple\mbox{-}Windowing module generates three clinically meaningful CT windows, including default, abdomen soft\mbox{-}tissue, and spine soft\mbox{-}tissue, and concatenates them as channels. This gives the model multi\mbox{-}contrast evidence for lesions and boundaries. The encoder then converts them into tokens with a patch embedding layer and extracts features through stacked VSS blocks and patch merging. This hierarchy captures both local textures and long\mbox{-}range context that are important for head–and–neck anatomy.

On every skip path, we place a Multi\mbox{-}Frequency Enhancement (MFE) module that splits the feature map into four branches. Three branches apply view\mbox{-}specific 2D Discrete Fourier Transform filtering to emphasize directional and scale\mbox{-}aware frequency cues, and one uses depthwise convolution to retain fine spatial details. The decoder mirrors the encoder with patch expanding and VSS blocks, fusing each stage with the corresponding MFE\mbox{-}enhanced skip. A final $1{\times}1$ convolution produces the output segmentation logits at full resolution.

\vspace{-3mm}
\subsection{Windowing-Enhanced Module}
\label{subsec:windowing_module}

\begin{figure*}[t]
  \centering
    \includegraphics[width=0.6\textwidth,trim={0cm 0cm 0cm 0cm}, clip]{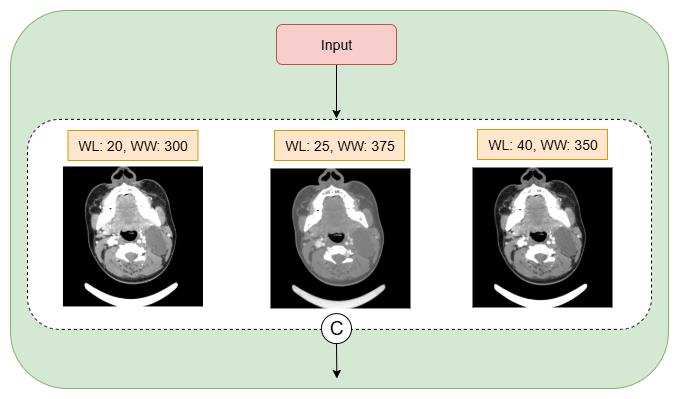}
    \vspace{-5 mm}
    \caption{Overview of Window-Enhanced Module}
    \label{fig:window}
    \vspace{-6 mm}
\end{figure*}
 In CT scan reading, radiologists select WL near the average attenuation of the target tissue and adjust WW to balance local contrast and global context \cite{xue2012window}. This step is necessary because CT values span a much wider dynamic range than both displays and human vision. Mimicking this progress exposes tissue boundaries and lesion locations that would otherwise be visually compressed. 

We instantiate three clinically meaningful windows for head--and--neck CT, including Default (WL/WW$=25/375$), Abdomen soft-tissue ($40/350$), and Spine soft-tissue ($20/300$) \cite{kann2018pretreatment,noda2018optimal} as Fig.~\ref{fig:window}. These settings restrict the effective input to HU ranges that cover muscles, glands, fat, pathologic soft tissue, and adjacent osseous structures that define lesion boundaries.
For each window $i\in\{1,2,3\}$ with level $L_i$ and width $W_i$, define
\begin{equation}\label{eq:minmax}
\mathrm{hu\_min}_i \;=\; L_i-\frac{W_i}{2}, \qquad
\mathrm{hu\_max}_i \;=\; L_i+\frac{W_i}{2},
\end{equation}
where $\mathrm{hu\_min}_i$ and $\mathrm{hu\_max}_i$ are the minimum and maximum HU values for the HU range of the customized window. 
Given the CT image in Hounsfield units $I_{\mathrm{HU}}\in\mathbb{R}^{H\times W}$, we first perform clipping as in Eq.~\eqref{eq:clip}, and then apply min-max normalize to $[0,1]$ as in Eq.~\eqref{eq:normalize}.

\begin{equation}\label{eq:clip}
I_i \;=\; \operatorname{clip}\!\big(I_{\mathrm{HU}},\, \mathrm{hu\_min}_i,\, \mathrm{hu\_max}_i\big),
\end{equation}
this equation clips values below $L-\frac{W}{2}$ to uniformly dark, and above $L+\frac{W}{2}$ to uniformly bright.

\begin{equation}\label{eq:normalize}
\tilde I_i \;=\; \frac{I_i - \mathrm{hu\_min}_i}{\mathrm{hu\_max}_i - \mathrm{hu\_min}_i} \;\in\; [0,1],
\end{equation}
where $\tilde I_i$ linearly maps intensities in $[\,L-\frac{W}{2},\,L+\frac{W}{2}\,]$ to $[0,1]$. 

Head--and--neck lesions exhibit heterogeneous density and often lie next to bone, air, and contrast-enhanced vessels. A single window cannot capture both low-contrast lesion interiors and high-contrast interfaces. The Abdomen soft-tissue window enhances parenchyma and lesion conspicuity; the Spine soft-tissue window preserves edges near bone--soft-tissue junctions; the Default window provides a balanced view when contrast timing or scanner protocols vary. Using all three windows, therefore, recovers complementary cues, such as fine boundaries, internal heterogeneity, and stable context, across a wide HU span \cite{xue2012window}.

From each slice, we generate three mapped images and concatenate them along the channel dimension:
\begin{equation}\label{eq:concate}
X \;=\; \mathrm{concat}\!\big[\tilde I_{1},\, \tilde I_{2},\, \tilde I_{3}\big] \;\in\; \mathbb{R}^{H\times W\times 3},
\end{equation}
This parameter-free fusion preserves the full dynamic content of each window rather than forcing an early average. The patch-embedding layer then projects $X$ into tokens for the downstream VSS blocks. Concatenation lets the network learn cross-window correspondences. For example, edges emphasized in the spine window align with subtle gradients in the soft-tissue window. It strengthens boundary localization, improves detection of small or low-contrast foci, and increases robustness to protocol and contrast variability \cite{kann2018pretreatment}.

\vspace{-2mm} 
\subsection{Multi-Frequency Enhancement Module}
\label{subsec:mfe}

\begin{figure*}[t]
  \centering
    \includegraphics[width=\textwidth,trim={0cm 0cm 0cm 0cm}, clip]{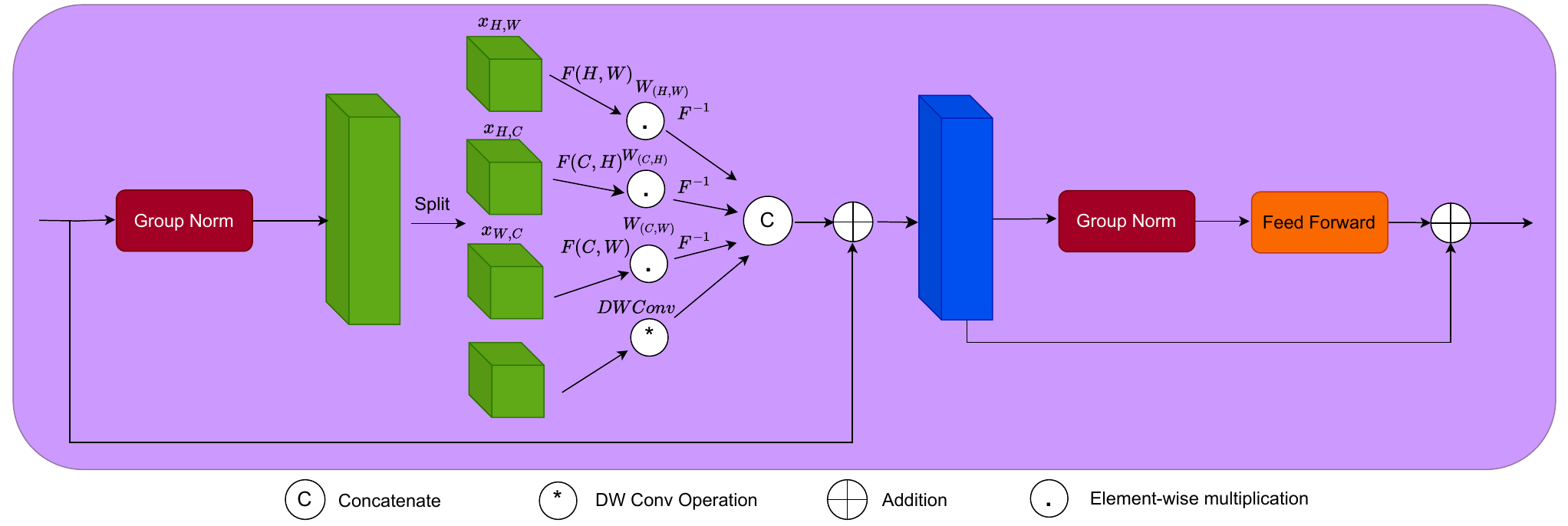}
    \vspace{-7 mm}
    \caption{Overview of Multi-Frequency Enhancement (MFE) Module}
    \label{fig:mfe}
    \vspace{-7 mm}
\end{figure*}

Besides spatial information, frequency information is critical for delineating small boundaries and textured patterns in medical images. Lesion edges, vessels, bone--soft tissue interfaces, and noise occupy different frequency bands and directions. Modeling them in the frequency domain makes these differences clearer and more separable. Our MFE module (as Fig.~\ref{fig:mfe}) is inserted on the skip connections of the WEMF model so that multi\mbox{-}frequency cues from high\mbox{-}resolution features are exposed to the decoder without losing the original spatial detail.

Given a feature map $X\!\in\!\mathbb{R}^{H\times W\times C}$ from each encoder block, we first apply Group Normalization to $X$, then partition its channels into four groups and process them in parallel. One branch captures global frequency information along the spatial plane ($H{\times}W$), two branches capture frequency along channel–spatial planes ($C{\times}W$ and $C{\times}H$), and one branch extracts local spatial detail using depthwise convolution (DWConv). 

\begin{equation}
\big(x_1,x_2,x_3,x_4\big) \;=\; \mathrm{Split}(\mathrm{Norm}(X)),
\label{eq:mfe-split}
\end{equation}

Let $\mathrm{F}_{(a,b)}$ and $\mathrm{F}_{(a,b)}^{-1}$ denote the 2D Discrete Fourier Transform (DFT) and Inverse Discrete Fourier Transform (IDFT) taken over axes $(a,b)\!\in\!\{(H,W), (C,W), (C,H)\}$. $W_{(a,b)}$ be a learnable complex frequency weight with the same shape as the transformed spectrum. The first three branches perform view\mbox{-}specific frequency filtering:
\begin{align}
\hat{x}_1 &= \mathrm{F}_{(H,W)}(x_1), \quad
x_1' \;=\; \mathrm{F}_{(H,W)}^{-1}\!\big(\hat{x}_1 \odot W_{(H,W)}\big), \label{eq:mfe-hw}\\
\hat{x}_2 &= \mathrm{F}_{(C,W)}(x_2), \quad
x_2' \;=\; \mathrm{F}_{(C,W)}^{-1}\!\big(\hat{x}_2 \odot W_{(C,W)}\big), \label{eq:mfe-cw}\\
\hat{x}_3 &= \mathrm{F}_{(C,H)}(x_3), \quad
x_3' \;=\; \mathrm{F}_{(C,H)}^{-1}\!\big(\hat{x}_3 \odot W_{(C,H)}\big), \label{eq:mfe-ch}
\end{align}
where $\odot$ denotes element\mbox{-}wise multiplication in the frequency domain. The fourth branch extracts local features for spatial detail:
\begin{equation}
x_4' \;=\; \mathrm{DWConv}(x_4),
\label{eq:mfe-dw}
\end{equation}
After that, the four outputs are concatenated and merged with the original skip features by a residual addition. 
This is important because head–and–neck structures are anisotropic. Different views emphasize different edge orientations and texture periodicities. Additionally, processing smaller channel groups allows for diverse frequency responses.
\begin{equation}
Z \;=\; \mathrm{Concat}\!\big[x_1',\,x_2',\,x_3',\,x_4'\big] \;+\; X,
\label{eq:mfe-out}
\end{equation}

Channel\mbox{-}wise concatenation preserves the full information from each view instead of averaging it early. This yields a richer representation that jointly encodes global low\mbox{-}frequency layout, directional high\mbox{-}frequency edges, inter\mbox{-}channel texture cues, and fine local details from DWConv. 
The addition with $X$ in Eq.~\eqref{eq:mfe-out} acts as an identity shortcut preserving the original skip information, so basic anatomical structure is not lost and stabilizes optimization and prevents over\mbox{-}filtering. 

After the residual fusion in Eq.~\eqref{eq:mfe-out}, we apply Group Normalization, a pointwise feed\mbox{-}forward mixing, and followed by a residual connection:
\begin{equation}
Y \;=\; \mathrm{FeedForward(\mathrm{Norm}(Z)}) \;+\; Z,
\label{eq:mfe-norm}
\end{equation}
This stage stabilizes feature statistics across the four branches and mixes information across channels so that frequency cues (from the three DFT views) and local cues (from DWConv) interact effectively before entering the decoder. The residual addition in Eq.~\eqref{eq:mfe-norm} preserves the original skip content for reliable anatomy while allowing the normalized and reweighted response to act as a controlled enhancement. In practice, this block incurs negligible overhead yet improves boundary sharpness and texture discrimination by aligning the multi\mbox{-}view responses in a common, well\mbox{-}conditioned space.

\vspace{-5mm}
\section{Experiments}

\vspace{-2mm}
\subsection{Implementation details}
Our dataset is split into training, validation, and testing sets with 2,931/278/570 slices, respectively.
We evaluated a set of segmentation architectures that represent the main deep-learning families used in medical imaging. We chose the CNN group, which includes U-Net \cite{ronneberger2015u} and U-Net++ \cite{zhou2018unet++}, and the Transformer group, which encompasses TransUNet \cite{chen2021transunet} and UNETR \cite{hatamizadeh2022unetr}. In addition, Mamba-based models, such as U-Mamba \cite{ma2024u}, Swin-UMamba \cite{liu2024swin}, and VM-UNet \cite{ruan2024vm}, were used for evaluation in our dataset.

We use a single NVIDIA RTX A5000 GPU with 24 GB for all experiments. Our model utilizes the AdamW optimizer with an initial learning rate of 0.0001 and employs a CosineAnnealingLR scheduler with \(T_{\text{max}} = 100\). Training ran for 300 epochs with a batch size of 16. 
We report both efficiency and accuracy. Accuracy on the MasHeNe dataset is evaluated using the Dice Similarity Coefficient (DSC), Intersection over Union (IoU), the 95th percentile Hausdorff Distance (HD95), Normalized Surface Distance (NSD), Accuracy, Recall, Specificity, and Precision. Efficiency is summarized by the number of parameters (millions) and floating-point operations (FLOPs). 

\vspace{-3mm}
\subsection{Quantitative Evaluation}
\begin{table}[!t]
\centering
\caption{Comparative performance of state-of-the-art models on the MasHeNe dataset }
\label{tab:results}
\vspace{-3mm}
\resizebox{\linewidth}{!}{
\begin{tabular}{l|cc|cccc|cccc}
\hline
\textbf{Model}       & \textbf{Param$\downarrow$}    & \textbf{FLOPs$\downarrow$}  & \textbf{DSC(\%)$\uparrow$} & \textbf{IoU(\%)$\uparrow$} & \textbf{HD95(mm)$\downarrow$} & \textbf{NSD(\%)$\uparrow$} & \textbf{Accuracy(\%)$\uparrow$} & \textbf{Recall(\%)$\uparrow$}& \textbf{Specificity(\%)$\uparrow$}& \textbf{Precision(\%)$\uparrow$}\\ \hline
U-Net \cite{ronneberger2015u}           & 31.04 M & 54.67 G & 66.19 & 63.00 & 9.00 & 68.62 & 99.67 & 67.13 & 99.80 & 68.62 \\
UNet++ \cite{zhou2018unet++}            & \textbf{9.16} M & 34.87 G & 67.29 & 64.37 & 3.96 & 68.61 & 99.73 & 68.07 & 99.86 & 68.61 \\
TransUNet \cite{chen2021transunet}      & 105.32 M & 32.25 G & 58.17 & 56.08 & 5.79 & 61.35 & 99.65 & 57.76 & \textbf{99.89} & 61.35 \\
UNETR \cite{hatamizadeh2022unetr}       & 87.20 M & 25.60 G & 38.73 & 36.11 & 25.72 & 44.45 & 99.25 & 38.00 & 99.58 & 44.45 \\
U-Mamba \cite{ma2024u}                  & 25.37 M & 842.62 G & 63.48 & 60.45 & 6.10 & 68.21 & 99.67 & 62.32 & 99.87 & 68.21 \\
Swin-UMamba \cite{liu2024swin}          & 59.88 M & 43.80 G & 54.54 & 51.23 & 6.89 & 62.36 & 99.61 & 53.12 & 99.85 & 62.36 \\
VM-UNet \cite{ruan2024vm}               & 27.42 M & \textbf{4.12} G & 63.21 & 59.67 & 11.01 & 65.44 & 99.60 & 64.13 & 99.73 & 65.44 \\ 
\hline
WEMF (Ours)   & 72.32 M & 12.19 G  & \textbf{70.45} & \textbf{66.89} & \textbf{5.12} & \textbf{72.33} & \textbf{99.78} & \textbf{71.07} & 99.88 &  \textbf{72.33}\\
\hline
\end{tabular}}
\vspace{-5mm}
\end{table}

In Table \ref{tab:results}, WEMF achieves the best overall segmentation quality on MasHeNe, with the highest DSC (70.45\%), IoU (66.89\%), and NSD (72.33\%), while the HD95 (5.12 mm) is the lowest. It also attains the best precision (72.33\%) and recall (71.07\%), near-perfect accuracy (99.78\%) and specificity (99.88\%), indicating few false positives and sharper boundary localization. These gains are consistent with our design, which includes triple-window inputs improve tissue and lesion contrast, while the multi-frequency module strengthens edge and texture cues, yielding better overlap and boundary metrics.
In terms of efficiency, WEMF uses 72.32M parameters and 12.19G FLOPs—heavier than the plain VM-UNet baseline but substantially lighter than transformer-heavy competitors, such as UNet++, TransUNet, or UNETR. This places WEMF in a favorable accuracy–efficiency regime for practical use.

When evaluating the Tumor and Cyst class as Table \ref{tab:class}, WEMF achieves the best metrics for both classes. This confirms consistent gains across lesion types. For boundary distance, WEMF is competitive but not always the very best. Particularly,  UNet++ attains the lowest Tumor HD95 (6.01 mm) and VM-UNet slightly edges the lowest Cyst HD95 (1.69 mm) versus WEMF’s 1.90 mm. Overall, the results indicate that the window-enhanced module and multi-frequency skip fusion improve overlap and surface conformity broadly.
\label{sec:exp}
\begin{table}[t]
\centering
\caption{Comparative performance of state-of-the-art models on the Tumor and Cyst class of the MasHeNe dataset.}
\label{tab:class}
\vspace{-3mm}
\resizebox{\linewidth}{!}{
\begin{tabular}{l|cccc|cccc}
\hline
\multicolumn{1}{c|}{\multirow{2}{*}{\textbf{Model}}}   & \multicolumn{4}{c|}{\textbf{Tumor}}                                                                  & \multicolumn{4}{c}{\textbf{Cyst}}                                                                    \\ \cline{2-9} 
\multicolumn{1}{c|}{}                                  & \textbf{DSC(\%)$\uparrow$} & \textbf{IoU(\%)$\uparrow$} & \textbf{HD95(mm)$\downarrow$} & \textbf{NSD(\%)$\uparrow$} & \textbf{DSC(\%)$\uparrow$} & \textbf{IoU(\%)$\uparrow$} & \textbf{HD95(mm)$\downarrow$} & \textbf{NSD(\%)$\uparrow$} \\ \hline
U-Net \cite{ronneberger2015u}           & 55.37 & 52.50 & 14.77 & 58.72 & 77.02 & 73.51 & 3.23 & 78.52 \\
UNet++ \cite{zhou2018unet++}            & 57.21 & 54.84 & \textbf{6.01} & 59.08 & 77.38 & 73.91 & 1.91 & 78.14 \\
TransUNet \cite{chen2021transunet}      & 48.82 & 47.55 & 9.74 & 52.86 & 67.52 & 64.62 & 1.84 & 69.84 \\
UNETR \cite{hatamizadeh2022unetr}       & 22.19 & 20.07 & 45.59 & 25.48 & 55.27 & 52.14 & 5.85 & 63.43  \\
U-Mamba \cite{ma2024u}                  & 54.80 & 52.21 & 9.41 & 59.69 & 72.15 & 68.69 & 2.79 & 76.73 \\
Swin-UMamba \cite{liu2024swin}          & 44.75 & 41.49 & 10.36 & 50.00 & 64.33 & 60.96 & 3.42 & 74.73 \\
VM-UNet \cite{ruan2024vm}               & 50.52 & 47.17 & 20.33 & 50.43 & 75.90 & 72.17 & \textbf{1.69} & \textbf{80.45} \\ \hline
WEMF (Ours)                             & \textbf{62.91} & \textbf{59.64} & 8.35 & \textbf{65.28} & \textbf{77.99} & \textbf{74.13} & 1.90 & 79.38\\ \hline
\end{tabular}}
\vspace{-2mm}
\end{table}

\subsection{Qualitative Analysis}

\begin{figure*}[t]
 \centering
   \includegraphics[width=\textwidth,trim={0cm 0cm 0cm 0cm}, clip]{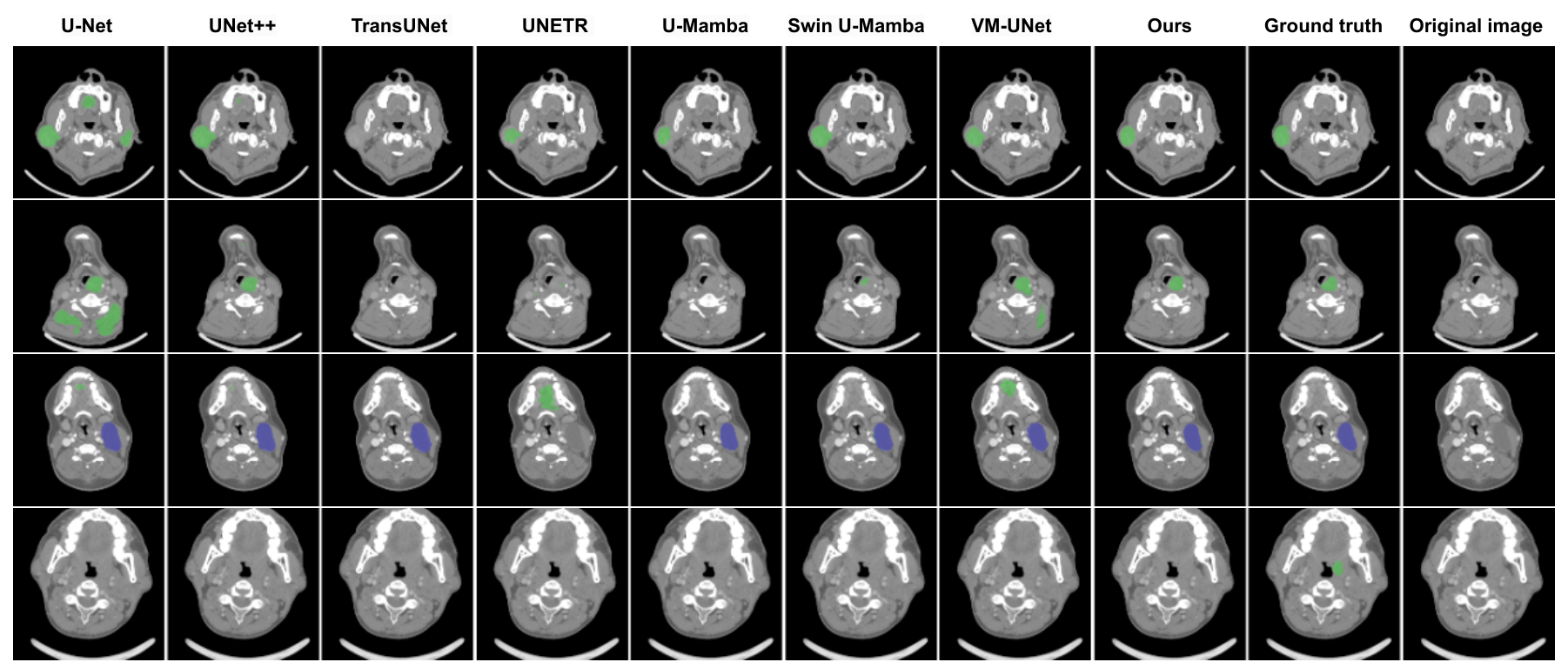}
   \vspace{-8 mm}
   \caption{Visualization of model performance on our MasHeNe dataset.}
   \label{fig:visualize}
   \vspace{-5 mm}
\end{figure*}
Figure~\ref{fig:visualize} presents qualitative results on the MasHeNe dataset. In the first and second rows, our method effectively suppresses false positives present in the predictions of U-Net, U-Net++, and VM-UNet, while also reducing the false negatives observed in TransUNet. For cystic lesions with clear margins and strong contrast (third row), most methods achieve satisfactory performance. In contrast, the final row presents challenging cases characterized by weak contrast and ambiguous boundaries, where most baseline models fail to accurately localize or segment the lesions. These difficult cases indicate the need for further improvement in future work.

\vspace{-2mm}
\subsection{Ablation study}
The proposed window-enhanced module delivers significant improvements over the baseline, with a gain of 5.96\% in DSC and 2.52\% in IoU, alongside a substantial reduction of 7.22 mm in HD95. These results confirm the utility of integrating multi-contrast information.
Further gains are realized by incorporating the MFE module into the skip connections, which attains a DSC of 68.15\%, an IoU of 64.95\%, and a state-of-the-art HD95 of 3.67 mm. This outcome underscores the effectiveness of frequency-domain cues in enhancing boundary delineation.
The synergistic integration of both modules produces the optimal result, achieving a peak DSC of 70.45\% and an IoU of 66.89\% with minimal compromise in HD95. This combination indicates that the modules capture complementary features.
\begin{table}[!t]
\centering
\caption{Ablation study for components in our WEMF model}
\vspace{-3 mm}
\label{tab:ablation}
\resizebox{\linewidth}{!}{
\begin{tabular}{cccc|cccc}
\hline
\multicolumn{3}{c|}{\textbf{Window}}  & \multicolumn{1}{c|}{\multirow{2}{*}{\textbf{MFE}}} & \multicolumn{1}{c}{\multirow{2}{*}{\textbf{DSC(\%)$\uparrow$}}} & \multicolumn{1}{c}{\multirow{2}{*}{\textbf{IoU(\%)$\uparrow$}}} & \multicolumn{1}{c}{\multirow{2}{*}{\textbf{HD95(mm)$\downarrow$}}} & \multicolumn{1}{c}{\multirow{2}{*}{\textbf{NSD(\%)$\uparrow$}}} \\ \cline{1-3}
\textbf{Default} & \textbf{Abdomen soft-tissue} & \multicolumn{1}{l|}{\textbf{Spine soft-tissue}} & \multicolumn{1}{c|}{} & \multicolumn{1}{c}{} & \multicolumn{1}{c}{} & \multicolumn{1}{c}{} & \multicolumn{1}{c}{} \\ \hline
$\checkmark$ & &                                        & & 63.21 & 59.67 & 11.01 & 65.44\\  
$\checkmark$ & $\checkmark$ & $\checkmark$              & & 69.17 & 65.73 & 3.79 & 71.57\\ 			
$\checkmark$ & &            & $\checkmark$                & 68.15 & 64.95& 3.67 & 72.78\\ \hline
$\checkmark$ & $\checkmark$ & $\checkmark$ & $\checkmark$ & 70.45 & 66.89&5.12 & 72.33\\ \hline
\end{tabular}}
\vspace{-5 mm}
\end{table}

\vspace{-3mm}
\subsection{Limitations and Future Perspectives}
The primary limitations of this study are the dataset scale and model performance on complex cases. The current dataset of 65 CECT cases, featuring only tumors and cysts, may not capture the full variability of head-and-neck lesions. To improve generalization, future works will expand the dataset with more cases, additional pathological categories (e.g., abscess, hematoma), and multi-center data. Furthermore, while WEMF-Net demonstrates improved overlap and boundary metrics, it still struggles with lesions exhibiting weak contrast and ill-defined margins. Subsequent research will therefore investigate advanced strategies, including boundary-aware losses, explicit small-object modeling, and adaptive post-processing, to enhance performance in addressing these challenges.
\vspace{-2mm}
\section{Conclusion}
\label{sec:conc}
We introduced MasHeNe, an initial benchmark for head-and-neck mass segmentation on contrast-enhanced CT. The dataset contains 65 cases with expert pixel-level masks for tumors and cysts, standardized splits, and common metrics to enable fair comparison across methods. This resource addresses a gap in current public corpora, which largely focus on malignancy and underrepresent other space-occupying lesions relevant to clinical care.
We also presented WEMF, a simple effective model combining triple-windowing enhancement with cross-frequency attention in a Mamba backbone. On MasHeNe, WEMF achieved the best performance among evaluated baselines, demonstrating stable behavior on a challenging task and highlighting the value of frequency-aware fusion with multi-window inputs.
In future work, we plan to expand MasHeNe to include additional lesion types and institutions, and release stronger multi-window, frequency-aware baselines to advance reproducible research in this domain.

\section*{Acknowledgment}
  This research is funded by Vietnam National University - Ho Chi Minh City (VNU-HCM) under grant number 36-2024-44-02.
%
%
%
\bibliographystyle{splncs04}
\bibliography{mybib}

\end{document}